\documentclass{article}

\usepackage{microtype}
\usepackage{graphicx}
\usepackage{subfigure}
\usepackage{booktabs} 

\usepackage[textsize=small]{todonotes}

\usepackage{hyperref}

\usepackage[accepted]{icml2021}

\icmltitlerunning{VEGN: variant effect prediction with graph neural network}

\begin{document}

\twocolumn[
\icmltitle{VEGN: Variant Effect Prediction with Graph Neural Networks}



\icmlsetsymbol{equal}{*}

\begin{icmlauthorlist}
\icmlauthor{Jun Cheng}{equal,nle}
\icmlauthor{Carolin Lawrence}{equal,nle}
\icmlauthor{Mathias Niepert}{nle}
\end{icmlauthorlist}

\icmlaffiliation{nle}{NEC Laboratories Europe, Germany}

\icmlcorrespondingauthor{Jun Cheng}{s6juncheng@gmail.com}

\icmlkeywords{Machine Learning, ICML}

\vskip 0.3in
]



\printAffiliationsAndNotice{\icmlEqualContribution} 

\begin{abstract}
Genetic mutations can cause disease by disrupting normal gene function. Identifying the disease-causing mutations from millions of genetic variants within an individual patient is a challenging problem. Computational methods which can prioritize disease-causing mutations have, therefore, enormous applications. It is well-known that genes function through a complex regulatory network. However, existing variant effect prediction models only consider a variant in isolation. In contrast, we propose VEGN, which models variant effect prediction using a graph neural network (GNN) that operates on a heterogeneous graph with genes and variants. 
The graph is created by assigning variants to genes and connecting genes with an gene-gene interaction network. In this context, we explore an approach where a gene-gene graph is given and another where VEGN learns the gene-gene graph and therefore operates both on given and learnt edges.
The graph neural network is trained to aggregate information between genes, and between genes and variants. Variants can exchange information via the genes they connect to. This approach improves the performance of existing state-of-the-art models.
\end{abstract}

\section{Introduction}
High-throughput sequencing approaches have revealed hundreds of millions of genetic variants across the human population. Each individual genome contains on average 4.1 million to 5 million variants from the human reference genome \citep{10002015global}. The vast majority of these variants are neutral and contribute to the genetic diversity of the human genome. This poses a challenge for computational tools to distinguish disease-causing variants from millions of neutral ones. The clinical application of genome sequencing in diagnosing disease is limited by the accuracy of current genome interpretation tools. As of today, we do not understand most of the relationships between human genetic variants and diseases~\citep{landrum2018clinvar}. Improving the performance of variant interpretation has a potentially  high-impact in clinical genetics.

Previous approaches of predicting variant disease consequences have been focused on predicting the functional consequences with optionally using evolutionary conservation signals \citep{rentzsch2019cadd, sundaram2018predicting, zhou2018deep, frazer2020large}. These methods tackle the variant effect prediction task by classifying variants into disease-causing (pathogenic) or benign. 

Despite the success, previous methods model variants in the classification tasks as independent observations. Such assumption might fail to capture important relations between variants. First, variants disrupting the same gene might lead to the same disease consequence. For instance, Spinal Muscular Atrophy can be caused by different mutations of the same SMN1 gene \citep{lefebvre1995identification}. Second, variants disrupting genes in the same functional group or pathway are also likely to have correlated disease consequences. The current annotation of disease variants is very sparse \citep{stenson2009human}. Considering the relatedness of variants in terms of disease-relevance may help annotating the remaining variants more precisely. Third, effect of variants may propagate through the gene regulatory network. A large survey of missense mutations across Mendelian disorders reviewed that most missense disease mutations do not affect protein folding or stability but the interaction with other molecules \citep{sahni2015widespread}. Examining \textit{de novo} mutations from Autism patients found an enrichment of missense variants on the interaction interface compared to healthy sibling controls \citep{chen2018interactome}. This evidence highlights the importance of considering variant effect in the context of a gene regulatory network.

\begin{figure*}[]
	\centerline{\includegraphics[width=0.98\textwidth]{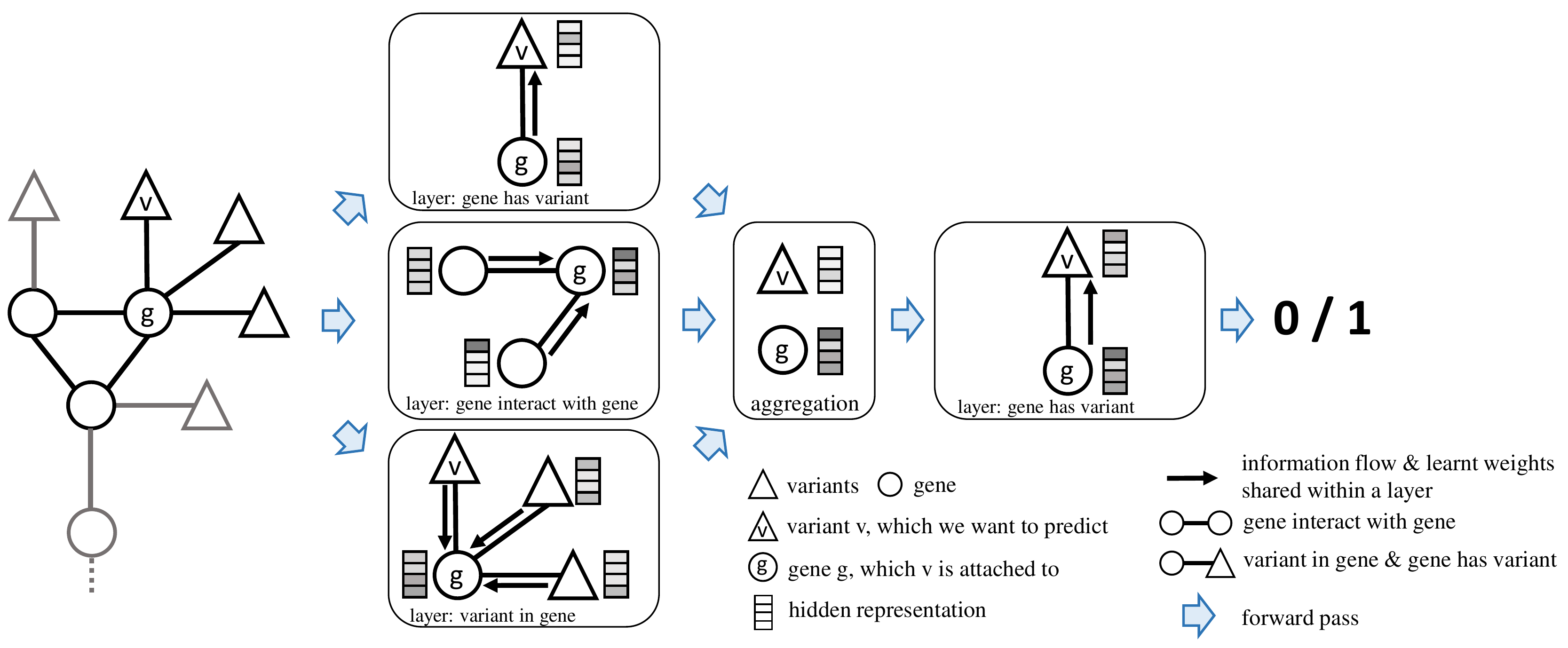}}
	\caption{Overview of the graph neural network which classifies variants (denoted by triangles) as benign (0) or pathogenic (1). Variants are associated with a gene (denoted by circles) and a gene-gene network is learnt. For the three different edge types (``gene \textit{has} variant'', ``gene \textit{interacts} with gene'' and ``variant \textit{in} gene''), separate GNN layers are instantiated and learnt. The hidden representations of each layer are aggregated, e.g. by sum. Finally, there is one further GNN layer where information flows from the gene to the variant. Based on this update, a final classification layer  determines the likelihood of a variant being benign or pathogenic. During training, the true label of a variant $v$ is observed and weights can be updated via a loss function and backpropagation. During test time, new variants can be added to the graph via the gene they attach to. Based on the features associated with the variant, the learnt weights can be applied in a forward pass to derive a prediction.}\label{fig:overview}
\end{figure*}

Graph neural networks are a powerful method to propagate and aggregate information through a graph. In this work, we propose a graph neural network approach that operates on a heterogeneous graph with genes and variants. The graph is created by assigning variants to genes based on the genome coordinates and connecting genes in a gene-gene interaction network. Variant effect prediction with such a GNN allows variants to draw information from the genes they connect to as well as other variants that connect to the same genes.
For the gene-gene network, we explore two approaches: (1) the a gene-gene graph that is given by a domain expert, (2) the gene-gene network is automatically learnt. 
In the first approach, the GNN is restricted to operate on given gene-gene edges. This issue is alleviated by the second approach, which allows us to find novel gene-gene connections.
With the latter approach we are, to the best of our knowledge, the first to combine both given and learnt (latent) edges in a heterogeneous (multi-edge type) graph. We show that both methods improve the prediction accuracy compared to previous state of the art. Additionally, it allows experts to interpret the prediction by inspecting which variants and genes had a large effect on a prediction.

\section{Method}
\subsection{Graph Neural Networks (GNNs)}
In a graph neural network, messages can be passed between nodes via edges that connect them. Concretely, we assume we are given a graph $\mathcal{G} = (\mathcal{N}, \mathcal{E})$, where $\mathcal{N}$ denotes the set of nodes and $\mathcal{E}$ the set of edges, where an edge $e_{ij} \in \mathcal{E}$ holds between the two nodes $n_i, n_j \in \mathcal{N}$. Furthermore, let $\mathcal{A}(i)$ denote the set of nodes that have an edge with $n_i$. Each node $n_i$ is associated with a feature $\mathbf{x}_i$, which is the input to the GNN. Hidden node representations can then be computed via

\begin{equation}\label{eq:node}
	\mathbf{x}_i^{(k)} = h^{(k)}\left(\mathbf{x}_i^{(k-1)}, \texttt{AGG}_{j \in \mathcal{A}(i)}f^{(k)}(\mathbf{x}_j^{(k-1)})\right),
\end{equation}

where $k$ refers to the $k$-th layer, $h^{(k)}$ and $f^{(k)}$ are differentiable functions parametrized by weights $\mathbf{w}$, $\texttt{AGG}$ is an aggregation function, such as the sum or mean, and with initial layer $\mathbf{x}_i^{(0)} = \mathbf{x}_i$. 
Different instantiations of Eq. \ref{eq:node} are possible and we discuss two possible options in Section \ref{sec:vep}. 

At the final layer $K$, $\mathbf{x}_i^{(K)}$ can be used for node classification. Given a training set where some nodes $n_i$ are associated with a classification label $c_i$, the weights of the GNN can be learnt using (stochastic) gradient descent and a loss function that determines the loss between model prediction and classification label.

The above framework can be extended to a heterogeneous graph, where edges of different types exist: for each edge type $\mathcal{E}_t$, we compute at each layer an edge specific representation $\mathbf{x}_{i,t}^{(k)}$ (using different instantiations of Eq. \ref{eq:node}), which can then be combined into an overall representation via an additional aggregation function (here a sum):

\begin{equation}\label{eq:type}
	\mathbf{x}_i^{(k)} = \texttt{SUM}_{\forall t}(\mathbf{x}_{i,t}^{(k)}).
\end{equation}


\subsection{Variant Effect Prediction with GNNs}\label{sec:vep}
For variant effect prediction with GNNs, we define the set of nodes as the combined set of variants and genes. 
We assume the set of genes to be fixed, whereas the set of variants is fixed during training but new ones can be added at prediction time.  For each variant, we assume a given feature vector $v_i$, which can for example be a prediction from non-graph-based models. For each gene the feature vector $g_i$ is randomly initialized and learnt during training.

Edges between nodes can occur either between two genes $(g_i, g_j)$ or between a gene and a variant $(g_i, v_j)$. Based on this, we define three different edge types, $\{has, in, interact\}$ where the edge sets are $\mathcal{E}_{has}$: all edges from a gene to a variant, $\mathcal{E}_{in}$: all edges from variants to genes, $\mathcal{E}_{interact}$ all edges between genes. For each edge type a different GNN is learnt using Eq. \ref{eq:node} and representations are combined via Eq. \ref{eq:type}.

To train a GNN, the final layer utilizes a sigmoid activation function to map a variant $v_i$ to a value $y_i \in [0, 1]$. For each variant $v_i$ in the training data, we can utilize their associated classification label $c_i = \{0, 1\}$ to compute a binary cross entropy loss. A graphical overview of the entire architecture is given in Figure \ref{fig:overview}. For instantiating Eq. \ref{eq:node}, we explore two different approaches, which we discuss next.



\subsubsection{Graph Construction based on Domain Knowledge}
\begin{figure}
	\centering
	\includegraphics[width=0.8\linewidth]{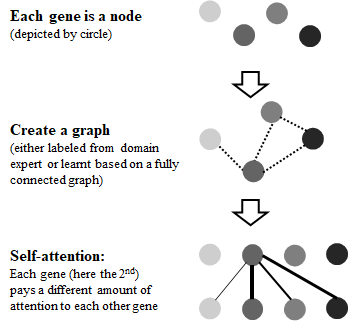}
	\caption{\label{fig:self_attention} The strength of gene-gene interactions can be learnt by using self-attention. Each gene is represented as a node and a graph is created. The self-attention layer learns for each gene how important each other gene is. If a gene-gene graph is given, attention can only flow between genes that are connected by an edge; for example in the graph above, no attention can flow between the darkest and the lightest node as there is no edge connecting them.}
\end{figure}

In the first approach, we assume that a gene-gene graph is given, where a domain expert determined which genes interact with which other genes. Here, we use the \textsc{Genemania} graph \citep{warde2010genemania}. 
Based on this given gene-gene graph, we can instantiate Eq. \ref{eq:node} for the edge type $\mathcal{E}_{interact}$ with a graph attention network (GAT) \cite{gat}. A GAT learns end-to-end in a self-attention layer how much attention a node should pay to each of its neighbours; higher attention indicates that this neighbour is more important for the current node (see also Figure \ref{fig:self_attention} as an example of how self-attention between genes can be computed). 

For each variant we create edges to the closest genes, which we can determine with the VEP tool \citep{mclaren2016ensembl}. Based on this, we also use GATs for the edge types $\mathcal{E}_{has}$ and $\mathcal{E}_{in}$. The input features for variants are predictions of PrimateAI model \citep{sundaram2018predicting}, which is the state-of-the-art model to predict effects of missense variants, but does not model the relateness between variants and genes. 

\subsubsection{Learning a Gene-Gene Graph}
In the previous approach, attention can only flow between two nodes if they have an edge connecting them. Consequently, if a gene-gene connection is important but not present in the provided gene-gene graph, then these two genes cannot exchange information directly. For this reason, we explore, in the second approach, whether it is possible to directly learn the gene-gene graph from the data with a supervised approach. 

To be able to learn a gene-gene graph, we start by creating a fully connected graph, where each gene is connected to every other gene. We would now like to compute self-attention for each gene to every other gene. However, the standard self-attention with the GAT framework becomes intractable for such a large fully connected graph: If a gene is connected to every other gene and there are $|G|$ genes, the computational complexity is $O(|G|^2d)$ (where $d$ is the chosen dimension of the hidden representation) and this quadratic dependence on $|G|$ becomes computationally too expensive for our graph with $|G|=19,551$ genes.

To reduce the computational complexity, we instead use the self-attention proposed by the Performer \cite{performer}. The performer offers an approximation to the standard self-attention mechanism, resulting in a lower computational complexity of $O(|G|d^2\log(d))$. The linear dependence on $|G|$ makes it computationally feasible to learn the gene-gene interactions end-to-end.

\section{Experiments}
With our experiments we would like to verify whether variant effect prediction can be improved by aggregating information through a gene-variant graph with a graph neural network. We focused on classifying missense variants. In particular, we are interested in whether adding information with graph can improve the performance of existing variant effect prediction tools. 

\subsection{Training Details}


Models are trained for 50 epochs and with batch size 20,480. The optimizer is \textsc{Adam} \cite{adam} with an initial learning rate of 0.01 that is reduced by 10-fold if no improvement has been observed on the evaluation set over 2 epochs.  For the GAT, we use two self-attention heads and an embedding size of 32 and 64 for the genes and variants, respectively. For the Performer, we use 2 self-attention heads and concatenate 3 such layers; dropout is set to 0.2 and the embedding size is 32 for both genes and variants.


\subsection{Datasets}
\textbf{\textit{gnomAD}} Variants data from the Genome Aggregation Database (gnomAD) v3.01 were filtered for single-nucleotide missense variants with minor allele frequency larger than 1\% across the population, resulting in 17,992 variants.

\noindent
\textbf{\textit{HGMD}} Variants from the human gene mutation database (HGMD, professional version, accessed Nov. 2020) were filtered for 138,689 missense disease-causing variants (DM).

\noindent
\textbf{\textit{ClinVar}} 15,943 annotated pathogenic missense variants from the ClinVar database (access date 14.11.2020).

\noindent
\textbf{\textit{Primate}} PrimateAI \citep{sundaram2018predicting} collected benign missense variants from human (83,546 variants) and 6 other primate species (301,690 variants). Benign variants are defined as common variants (minor allele frequency $>$ 0.1\% in corresponding species). Another 68,258,623 all possible missense variants in human, with variants observed in from ExAC and start/stop variants excluded, were collected and referred as the "unlabeled" set of variants. Since the unlabeled variants are extremely rare in human population, they are considered to be eliminated by selection pressure, and therefore are used as proxy for pathogenic variants during training. We converted the hg19 coordinates of this dataset to hg38 with the Picard \textit{LiftoverVcf} tool \citep{Picard2019toolkit}.

\noindent
\textbf{\textit{Test Set}} We combine the pathogenic variants from the \textit{HGMD} and the \textit{ClinVar} as the positive samples in the \textit{test set}. We matched the number of pathogenic variants by randomly selecting common missense variants from \textit{gnomAD}. We refer the common variants from gnomAD as benign.

\noindent
\textbf{\textit{Training \& Evaluation Set}} We train and evaluate our models with the \textit{Primate} data. 80\% of the variants were randomly selected as the training set and the remaining as the evaluation set. 

\subsection{Results}

We report results in terms of area under the receiver operating characteristic curve (auROC) in Table \ref{Tab:01}.
We compared our two models, VEGN with gene-gene graph (G-G) given or learnt to the PrimateAI model. Both VEGN models outperform the PrimateAI model. As VEGN uses the PrimateAI predictions as input features for the variants, the improvements can be directly attributed to using a GNN on the heterogeneous gene and variant graph. In the future, we plan to explore if further improvements can be achieved by collecting additional variant features.

Comparing the two different VEGN models, we find that the model which learns the gene-gene graph slightly outperforms the model where the graph is given. This is an encouraging result as it indicates that the model is able to automatically discover relevant gene-gene interactions without any human annotation necessary. Future work in this direction might even help us to uncover biologically meaningful but as of yet unknown gene-gene interactions.

Finally, another advantage of our model compared to PrimateAI is the ability for domain experts to directly inspect which genes and variants had a high influence on the prediction of a particular variant. Such an analysis can for example be done by inspecting the learned attention weights \cite{abnar2020}.


\begin{table}[H]
\begin{center}
{\begin{tabular}{@{}ll@{}} \toprule
Method         & auROC \\ \midrule
PrimateAI          & 0.8162         \\
VEGN G-G given & 0.8267  \\
VEGN G-G learnt        & \textbf{0.8291}       \\ \toprule
\end{tabular}}{}
\caption{Comparing VEGN with gene-gene graph (G-G) given or G-G learnt to the PrimateAI model. \label{Tab:01}}
\end{center}
\end{table}

\section*{Discussion}
Variant effect prediction is a long-standing challenge in biology. 
We show that integrating and aggregating information through a heterogeneous gene-gene and gene-variant graph with a graph neural network (GNN) can improve variant effect prediction. This approach is distinct from existing approaches where variants are classified in isolation. A GNN model has the advantage of considering the relatedness of variants as opposed to considering variants as independent observations in previous models. 
The performance improvement in this work is likely limited by the fact that we simply used PrimateAI scores as variant features. Future work can address this shortcoming by exploring other options for variant features, for example by using learnt representations of a deep neural network that can be trained jointly with the GNN end-to-end. 

\bibliography{document}

\begin{thebibliography}{17}
\providecommand{\natexlab}[1]{#1}
\providecommand{\url}[1]{\texttt{#1}}
\expandafter\ifx\csname urlstyle\endcsname\relax
  \providecommand{\doi}[1]{doi: #1}\else
  \providecommand{\doi}{doi: \begingroup \urlstyle{rm}\Url}\fi

\bibitem[Pic(2019)]{Picard2019toolkit}
Picard toolkit.
\newblock http://broadinstitute.github.io/picard/, 2019.

\bibitem[Abnar \& Zuidema(2020)Abnar and Zuidema]{abnar2020}
Abnar, S. and Zuidema, W.
\newblock Quantifying attention flow in transformers.
\newblock In \emph{Proceedings of the 58th Annual Meeting of the Association
  for Computational Linguistics}, pp.\  4190--4197, Online, July 2020.
  Association for Computational Linguistics.

\bibitem[Chen et~al.(2018)Chen, Fragoza, Klei, Liu, Wang, Roeder, Devlin, and
  Yu]{chen2018interactome}
Chen, S., Fragoza, R., Klei, L., Liu, Y., Wang, J., Roeder, K., Devlin, B., and
  Yu, H.
\newblock An interactome perturbation framework prioritizes damaging missense
  mutations for developmental disorders.
\newblock \emph{Nature Genetics}, 50\penalty0 (7):\penalty0 1032--1040, 2018.

\bibitem[Choromanski et~al.(2020)Choromanski, Likhosherstov, Dohan, Song, Gane,
  Sarlos, Hawkins, Davis, Mohiuddin, Kaiser, et~al.]{performer}
Choromanski, K., Likhosherstov, V., Dohan, D., Song, X., Gane, A., Sarlos, T.,
  Hawkins, P., Davis, J., Mohiuddin, A., Kaiser, L., et~al.
\newblock Rethinking attention with performers.
\newblock \emph{arXiv preprint arXiv:2009.14794}, 2020.

\bibitem[Consortium et~al.(2015)]{10002015global}
Consortium, . G.~P. et~al.
\newblock A global reference for human genetic variation.
\newblock \emph{Nature}, 526\penalty0 (7571):\penalty0 68, 2015.

\bibitem[Frazer et~al.(2020)Frazer, Notin, Dias, Gomez, Brock, Gal, and
  Marks]{frazer2020large}
Frazer, J., Notin, P., Dias, M., Gomez, A., Brock, K., Gal, Y., and Marks, D.
\newblock Large-scale clinical interpretation of genetic variants using
  evolutionary data and deep learning.
\newblock \emph{bioRxiv}, 2020.

\bibitem[Kingma \& Ba(2015)Kingma and Ba]{adam}
Kingma, D.~P. and Ba, J.
\newblock Adam: {A} method for stochastic optimization.
\newblock In \emph{3rd International Conference on Learning Representations,
  {ICLR}}, 2015.

\bibitem[Landrum \& Kattman(2018)Landrum and Kattman]{landrum2018clinvar}
Landrum, M.~J. and Kattman, B.~L.
\newblock Clinvar at five years: delivering on the promise.
\newblock \emph{Human mutation}, 39\penalty0 (11):\penalty0 1623--1630, 2018.

\bibitem[Lefebvre et~al.(1995)Lefebvre, B{\"u}rglen, Reboullet, Clermont,
  Burlet, Viollet, Benichou, Cruaud, Millasseau, Zeviani,
  et~al.]{lefebvre1995identification}
Lefebvre, S., B{\"u}rglen, L., Reboullet, S., Clermont, O., Burlet, P.,
  Viollet, L., Benichou, B., Cruaud, C., Millasseau, P., Zeviani, M., et~al.
\newblock Identification and characterization of a spinal muscular
  atrophy-determining gene.
\newblock \emph{Cell}, 80\penalty0 (1):\penalty0 155--165, 1995.

\bibitem[McLaren et~al.(2016)McLaren, Gil, Hunt, Riat, Ritchie, Thormann,
  Flicek, and Cunningham]{mclaren2016ensembl}
McLaren, W., Gil, L., Hunt, S.~E., Riat, H.~S., Ritchie, G.~R., Thormann, A.,
  Flicek, P., and Cunningham, F.
\newblock The ensembl variant effect predictor.
\newblock \emph{Genome Biology}, 17\penalty0 (1):\penalty0 1--14, 2016.

\bibitem[Rentzsch et~al.(2019)Rentzsch, Witten, Cooper, Shendure, and
  Kircher]{rentzsch2019cadd}
Rentzsch, P., Witten, D., Cooper, G.~M., Shendure, J., and Kircher, M.
\newblock Cadd: predicting the deleteriousness of variants throughout the human
  genome.
\newblock \emph{Nucleic Acids Research}, 47\penalty0 (D1):\penalty0 D886--D894,
  2019.

\bibitem[Sahni et~al.(2015)Sahni, Yi, Taipale, Bass, Coulombe-Huntington, Yang,
  Peng, Weile, Karras, Wang, et~al.]{sahni2015widespread}
Sahni, N., Yi, S., Taipale, M., Bass, J. I.~F., Coulombe-Huntington, J., Yang,
  F., Peng, J., Weile, J., Karras, G.~I., Wang, Y., et~al.
\newblock Widespread macromolecular interaction perturbations in human genetic
  disorders.
\newblock \emph{Cell}, 161\penalty0 (3):\penalty0 647--660, 2015.

\bibitem[Stenson et~al.(2009)Stenson, Mort, Ball, Howells, Phillips, Thomas,
  and Cooper]{stenson2009human}
Stenson, P.~D., Mort, M., Ball, E.~V., Howells, K., Phillips, A.~D., Thomas,
  N.~S., and Cooper, D.~N.
\newblock The human gene mutation database: 2008 update.
\newblock \emph{Genome medicine}, 1\penalty0 (1):\penalty0 1--6, 2009.

\bibitem[Sundaram et~al.(2018)Sundaram, Gao, Padigepati, McRae, Li, Kosmicki,
  Fritzilas, Hakenberg, Dutta, Shon, et~al.]{sundaram2018predicting}
Sundaram, L., Gao, H., Padigepati, S.~R., McRae, J.~F., Li, Y., Kosmicki,
  J.~A., Fritzilas, N., Hakenberg, J., Dutta, A., Shon, J., et~al.
\newblock Predicting the clinical impact of human mutation with deep neural
  networks.
\newblock \emph{Nature Genetics}, 50\penalty0 (8):\penalty0 1161--1170, 2018.

\bibitem[Veli{\v{c}}kovi{\'{c}} et~al.(2018)Veli{\v{c}}kovi{\'{c}}, Cucurull,
  Casanova, Romero, Li{\`{o}}, and Bengio]{gat}
Veli{\v{c}}kovi{\'{c}}, P., Cucurull, G., Casanova, A., Romero, A., Li{\`{o}},
  P., and Bengio, Y.
\newblock {Graph Attention Networks}.
\newblock \emph{International Conference on Learning Representations (ICLR)},
  2018.

\bibitem[Warde-Farley et~al.(2010)Warde-Farley, Donaldson, Comes, Zuberi,
  Badrawi, Chao, Franz, Grouios, Kazi, Lopes, et~al.]{warde2010genemania}
Warde-Farley, D., Donaldson, S.~L., Comes, O., Zuberi, K., Badrawi, R., Chao,
  P., Franz, M., Grouios, C., Kazi, F., Lopes, C.~T., et~al.
\newblock The genemania prediction server: biological network integration for
  gene prioritization and predicting gene function.
\newblock \emph{Nucleic Acids Research}, 38\penalty0 (suppl\_2):\penalty0
  W214--W220, 2010.

\bibitem[Zhou et~al.(2018)Zhou, Theesfeld, Yao, Chen, Wong, and
  Troyanskaya]{zhou2018deep}
Zhou, J., Theesfeld, C.~L., Yao, K., Chen, K.~M., Wong, A.~K., and Troyanskaya,
  O.~G.
\newblock Deep learning sequence-based ab initio prediction of variant effects
  on expression and disease risk.
\newblock \emph{Nature Genetics}, 50\penalty0 (8):\penalty0 1171--1179, 2018.

\end{thebibliography}
\bibliographystyle{icml2021}


\end{document}